\documentclass[9pt,conference]{IEEEtran}
\IEEEoverridecommandlockouts

\usepackage{cite}
\usepackage{amsmath,amssymb,amsfonts}
\usepackage{algorithmic}
\usepackage{graphicx}
\usepackage{textcomp}
\usepackage{xcolor}
\usepackage{balance}

\usepackage{booktabs}
\usepackage{hyperref}

\newcommand{\linebreakand}{%
  \end{@IEEEauthorhalign}
  \hfill\mbox{}\par
  \mbox{}\hfill
  \begin{@IEEEauthorhalign}
}
\def\BibTeX{{\rm B\kern-.05em{\sc i\kern-.025em b}\kern-.08em
    T\kern-.1667em\lower.7ex\hbox{E}\kern-.125emX}}
\begin{document}

\title{JELLY: Joint Emotion Recognition and Context Reasoning with LLMs for Conversational Speech Synthesis\\
\thanks{$^\dagger$Corresponding author. This work was partly supported by the Institute of Information \& Communications Technology Planning \& Evaluation (IITP) grant funded by the Korea government (MSIT) (No. RS-2019-II190079, Artificial Intelligence Graduate School Program (Korea University), No. RS-2021-II-212068, Artificial Intelligence Innovation Hub, No. RS-2024-00336673, AI Technology for Interactive Communication of Language Impaired Individuals, and No. RS-2024-00436857, Information Technology Research Center (ITRC) support program).}
}

\author{\IEEEauthorblockN{1\textsuperscript{st} Jun-Hyeok Cha}
\IEEEauthorblockA{\textit{Dept. of Artificial Intelligence} \\
\textit{Korea University}\\
Seoul, Korea \\
jh\_cha@korea.ac.kr}
\and
\IEEEauthorblockN{2\textsuperscript{nd} Seung-Bin Kim}
\IEEEauthorblockA{\textit{Dept. of Artificial Intelligence} \\
\textit{Korea University}\\
Seoul, Korea \\
sb-kim@korea.ac.kr}
\and
\IEEEauthorblockN{3\textsuperscript{rd} Hyung-Seok Oh}
\IEEEauthorblockA{\textit{Dept. of Artificial Intelligence} \\
\textit{Korea University}\\
Seoul, Korea \\
hs\_oh@korea.ac.kr}
\and
\IEEEauthorblockN{4\textsuperscript{th} Seong-Whan Lee$^{\dagger}$}
\IEEEauthorblockA{\textit{Dept. of Artificial Intelligence} \\ 
\textit{Korea University}\\
Seoul, Korea \\
sw.lee@korea.ac.kr}
}
\maketitle

\begin{abstract}Recently, there has been a growing demand for conversational speech synthesis (CSS) that generates more natural speech by considering the conversational context. 
To address this, we introduce JELLY, a novel CSS framework that integrates emotion recognition and context reasoning for generating appropriate speech in conversation by fine-tuning a large language model (LLM) with multiple partial LoRA modules.
We propose an Emotion-aware Q-former encoder, which enables the LLM to perceive emotions in speech. The encoder is trained to align speech emotions with text, utilizing datasets of emotional speech. The entire model is then fine-tuned with conversational speech data to infer emotional context for generating emotionally appropriate speech in conversation. 
Our experimental results demonstrate that JELLY excels in emotional context modeling, synthesizing speech that naturally aligns with conversation, while mitigating the scarcity of emotional conversational speech datasets. 
\end{abstract}


\section{Introduction}
Conversational speech synthesis (CSS) focuses on generating speech that has contextually appropriate prosody within a conversation. 
The primary distinction between CSS and text-to-speech (TTS) \cite{8461368, ren2021fastspeech, lee2022hierspeech, 10448291, 10445948} tasks is that conventional TTS does not consider prior conversation history. 
In contrast, CSS reflects the interactions between speakers within a conversation. 
Earlier CSS approaches \cite{9383460, 9747837, 10096905, 10446506, 10448356} explored various strategies for utilizing conversation history to generate natural speech in a conversational context. 
The GRU-based approach \cite{9383460} focuses on textual information of dialogue, extracting utterance-level semantic features to generate conversational speech. However, this approach has overlooked the information in speech, leading to limitations in fully capturing conversational context.

To address these limitations, \cite{9747837} enhances speaking style through graph-based multi-modal context modeling. 
M2CTTS \cite{10096905} proposes a method that leverages both textual and acoustic information, addressing coarse- and fine-grained features simultaneously within each domain. 
CONCSS \cite{10446506} introduces a contrastive learning-based method that generates context-sensitive representations for dialogue-appropriate prosody. 
Despite these advancements and progress in deep learning \cite{557671, jeong2019classification, lee2019towards, mane2020multi, won2020adaptive, lee2024periodwave}, fully capturing expressive conversational contexts like emotional speech \cite{9747098, 10517426, cho24_interspeech, oh2024durflex} remains challenging.
\begin{figure}[!t]
    \centering
    \centerline{\includegraphics[width=\columnwidth]{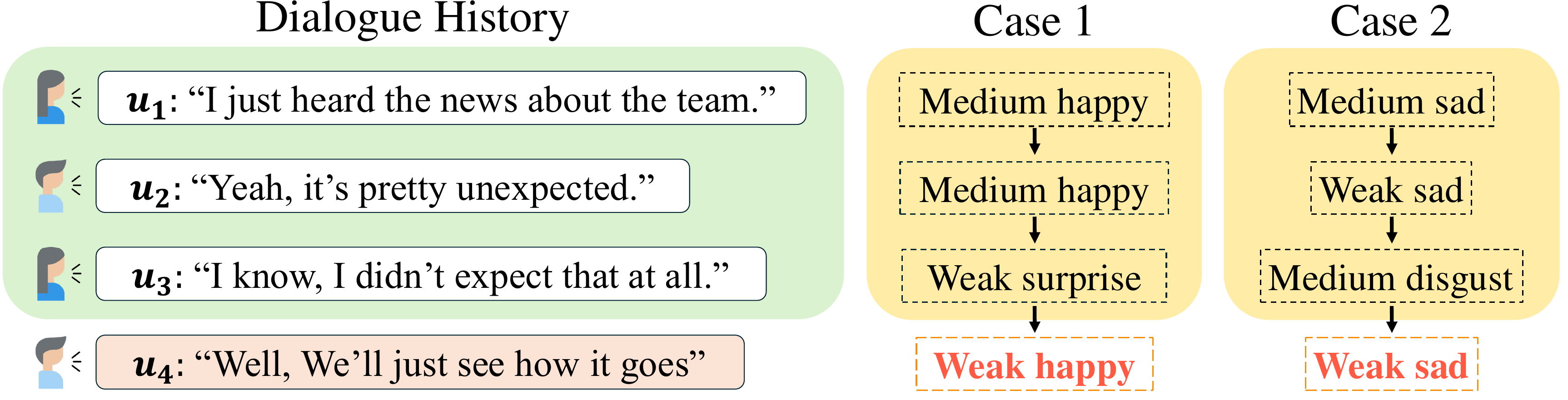}}
    \caption{Examples of different emotional contexts that can arise from the same conversation content.}
    \label{fig:example} 
    \vspace{-0.4cm}
\end{figure}
\begin{figure*}[htb!]
    \centering
        \includegraphics[width=\textwidth]{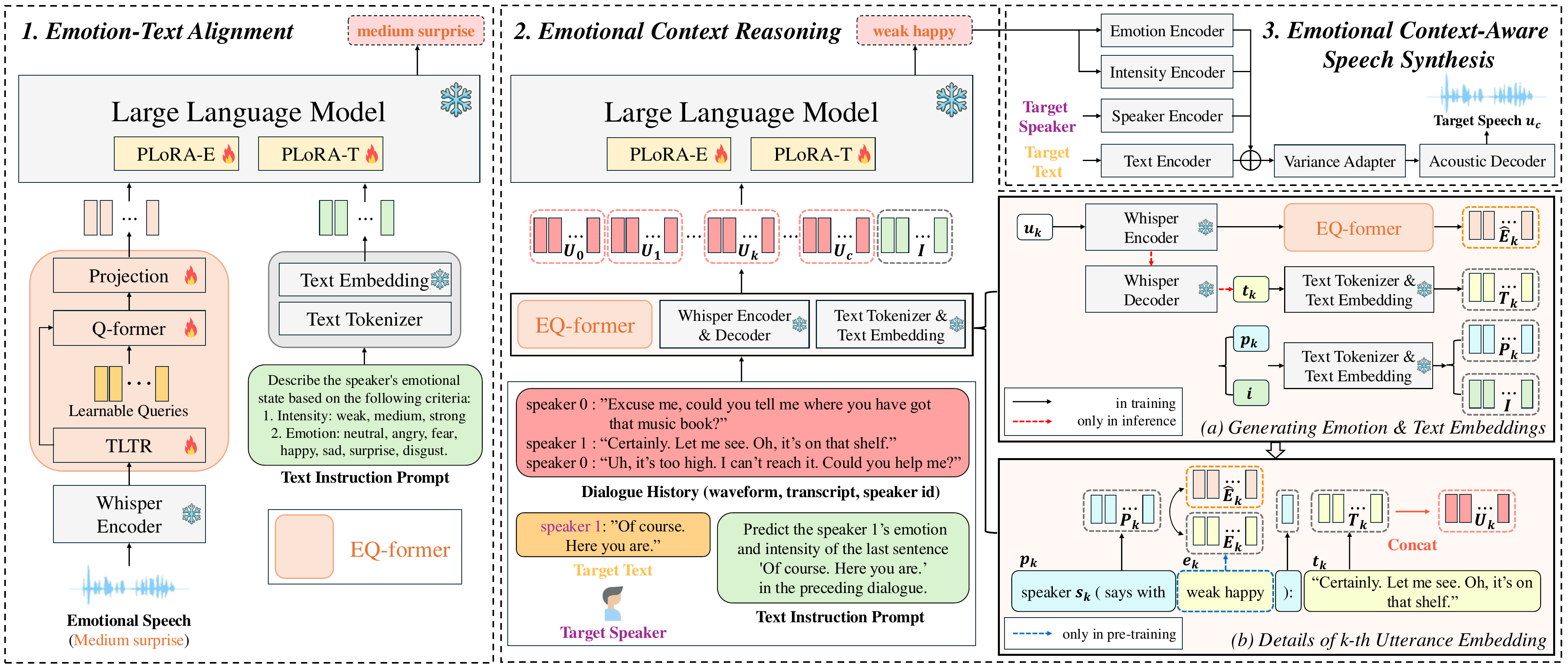}\vspace{-0.2cm}
    \caption{The overview of JELLY framework.}
    \vspace{-0.3cm}
    \label{fig:overview}
\end{figure*}

In human conversations, even when the literal content is identical, emotional context varies depending on the emotions conveyed by each speaker, as shown in Fig. \ref{fig:example}. 
Speech that is mismatched with the emotional context feels unnatural or awkward, emphasizing the importance of considering the emotional context in the CSS task. 
ECSS \cite{liu2024emotion} demonstrates the importance of considering emotion in the CSS task through a heterogeneous graph-based approach that predicts appropriate emotional expressions by using ground-truth emotion labels of each utterance within a conversation history. 

Still, understanding emotional context remains challenging in the CSS task, as it requires recognizing each utterance’s emotion and integrating it with the conversation content. 
Both steps are complex, even for humans. 
Moreover, the combined task of inferring emotional context and generating contextually appropriate speech requires high-quality emotional conversational datasets, which are scarce. 

To address these challenges, we introduce a novel framework, \textbf{JELLY} that incorporates \textbf{J}oint \textbf{E}motion recognition and context reasoning with a \textbf{L}arge \textbf{L}anguage model for conversational speech s\textbf{Y}nthesis. Our framework consists of three stages that enhance emotional context inference for CSS and alleviate the scarcity of emotional conversational speech datasets. 
To infer the emotional context, we leverage the reasoning ability of LLM through the partial LoRA (PLoRA) approach \cite{dong2024internlm, wang2024blsp}. 
To align speech emotions with text, we propose an Emotion-aware Q-former encoder (EQ-former). 
Integrating it with Whisper as a speech encoder enables the LLM to perceive emotional states from speech and infer the emotional context accordingly. 
By leveraging this speech encoder, JELLY infers the emotional context for CSS from speech alone during inference, making it more suitable to real-world scenarios.

Our contributions are summarized as follows:
\begin{itemize}
    \item We introduce JELLY, a novel CSS framework designed to synthesize emotionally appropriate speech in conversational contexts, mitigating the scarcity of emotional conversational data.
    \item Our method effectively infers emotional context solely from speech, unlike existing CSS models that require emotion labels or transcripts of dialogue history during inference.
    \item Experimental results exhibit that JELLY outperforms comparison models in emotional context reasoning and speech synthesis based on the conversational context. Audio samples and code are available at \url{https://jh-cha-prml.github.io/JELLY}.
\end{itemize}
\section{Methods}
\label{sec:methods}
\subsection{Emotion-aware Q-former Encoder}
\label{ssec:emotion_aware_qformer}
Comprehending the emotional context of a conversation requires recognizing the emotional state of the speaker in each utterance. 
Accordingly, we propose EQ-former as an emotion perception module for each utterance, as described in Fig. \ref{fig:overview}. 
The EQ-former consists of the Time and Layer-Wise \cite{gong23d_interspeech} Transformer (TLTR) for emotion feature extraction, the Querying Transformer (Q-former) \cite{pmlr-v202-li23q} for emotion alignment, and a projection layer. 

We adopt a Whisper encoder \cite{pmlr-v202-radford23a} and the TLTR module to extract emotional information from speech. 
The Whisper encoder representations encompass various types of speech information, such as paralinguistic cues \cite{gong23d_interspeech, 10389742}. 
We leverage these rich paralinguistic representations to capture emotion-related information. 
The Whisper encoder takes the waveform as input, and the intermediate representations from the 32 layers of the Whisper encoder are passed to the TLTR. 
We specifically apply the attention mechanism to enable the model to focus on the layers that contain more relevant emotional information through the TLTR. 

Additionally, we utilize the Q-former module to align emotion representations with text representations. 
Emotion features extracted by TLTR interact with a fixed set of learnable query embeddings through cross-attention layers in the Q-former module. 
The query tokens pass through a self-attention layer and then interact with the emotion features via the cross-attention layer, aligning the query outputs that hold emotional information with the text representation. 
This cross-modal alignment is crucial for enabling the LLM to reason about the emotional context by jointly understanding emotional information of each utterance and the content of the dialogue. 
Finally, we apply a linear layer to project the query outputs to match the embedding dimension of the LLM, ensuring that the projection layer is trainable. 
\subsection{LLM with Multiple PLoRA Modules}
\label{ssec:llm_multiple_plora}
To efficiently handle both emotion and text embeddings, we follow the partial LoRA (PLoRA) \cite{dong2024internlm, wang2024blsp} approach, which is designed to adapt the LLM's parameters to partial inputs. 
This approach preserves the LLM's integrity in encoding and generating text tokens while providing additional modeling capacity to address gaps between different modalities.

In our task, we fine-tune the LLM for emotional context reasoning, guiding its behavior depending on the input modality by employing multiple PLoRA modules. Specifically, PLoRA is applied to emotion embeddings (PLoRA-E) from the EQ-former and to text embeddings (PLoRA-T) from the LLM's tokenizer. 
By applying distinct PLoRA modules for each modality, we selectively adapt the LLM's parameters, effectively reducing the gap between emotion and text while fine-tuning the LLM for emotional context reasoning. 
\begin{table*}[!t]
    \centering
        \caption{Subjective and objective evaluation results. The N-DMOS and E-DMOS scores are presented with 95\% confidence intervals.} \vspace{0.1cm}
    \label{table_main}
    \vspace{-0.3cm}
    \resizebox{0.99\textwidth}{!}{
    \begin{tabular}{l|cc|c|cc|ccc|c|c}
        \toprule
        \textbf{Method} & \textbf{N-DMOS} $(\uparrow)$ & \textbf{E-DMOS} $(\uparrow)$ & \textbf{ECA} $(\uparrow)$ & \textbf{WER} $(\downarrow)$ & \textbf{PER} $(\downarrow)$ & $\textbf{RMSE}_{f0}$ $(\downarrow)$ & $\textbf{RMSE}_{p}$ $(\downarrow)$ & $\textbf{F1}_{v/uv}$ $(\uparrow)$ & \textbf{MCD} $(\downarrow)$ & \textbf{DDUR} $(\downarrow)$ \\
        \midrule    
            GT & 3.900 $\pm$ 0.045 & 4.063 $\pm$ 0.039 & 56.16 & 5.72 & 28.33 &- & - & - & - & - \\
            Vocoded \cite{NEURIPS2020_c5d73680} & 3.871 $\pm$ 0.043 & 4.013 $\pm$ 0.033 & 57.05 & 5.73 & 30.81 & 4.21 & 0.2610 & 0.8421 & 1.263 & 0.0572  \\
        \midrule 
            FastSpeech 2 \cite{ren2021fastspeech} & 3.788 $\pm$ 0.045 & 3.910 $\pm$ 0.043 & 56.38 & 6.06 & 29.50 & 23.65 & 0.4786 & 0.7280 & 3.313 & 0.2386  \\
            GRU-based \cite{9383460} & 3.792 $\pm$ 0.043 & 3.944 $\pm$ 0.034 & 56.38 & 5.91 & 29.80 & 22.21 & 0.4739 & 0.7277 & 3.487 & 0.2865  \\
            ECSS \cite{liu2024emotion} & 3.802 $\pm$ 0.045 & 3.914 $\pm$ 0.038 & 55.72 & \textbf{5.78} & 28.94 & \textbf{21.65} & 0.4765 & 0.7257 & 3.361 & 0.2525  \\
            \midrule 
            JELLY & \textbf{3.847} $\pm$ \textbf{0.042} & \textbf{3.987} $\pm$ \textbf{0.028} & \textbf{58.60} & 5.88 & \textbf{28.92} & 22.57 & \textbf{0.4733} & \textbf{0.7334} & 3.217 & \textbf{0.2157}  \\
            JELLY (speech-only) & 3.790 $\pm$ 0.043 & 3.983 $\pm$ 0.027 & 57.05 & 6.01 & 28.95 & 23.48 & 0.4759 & 0.7292 & \textbf{3.059} & 0.2194  \\

        \bottomrule
    \end{tabular}
    } 
\end{table*}
\subsection{Three-Stage Learning Pipeline}
\label{ssec:3_stage}
We designed a three-stage learning pipeline that enables 1) recognition of emotions in each utterance, 2) appropriate emotion inference based on both the perceived emotions and the conversation content, and 3) generation of speech that aligns with the inferred emotional context, mitigating the scarcity of emotional conversational datasets.  
\subsubsection{Emotion-Text Alignment} 
\label{sssec:stage01}
The goal of the first stage is to train the EQ-former to extract emotional information from an utterance and align the emotion features with the text embedding space. 
The emotion-text alignment stage is designed to enable the LLM to perceive the emotional states of each utterance in the dialogue history. 
The EQ-former is trained using the LLM's next-token prediction task to extract emotion representations. 
We adapt the LLM with multiple PLoRA modules to determine the emotional states of the target speech, using emotion embeddings from the EQ-former and text embeddings from the instruction prompt, as shown in the left panel of Fig. \ref{fig:overview}. 
The EQ-former and PLoRA modules are trained to align text with speech emotions, leveraging the more readily available emotional speech data. 
During training, the Whisper encoder and LLM are frozen.
\subsubsection{Emotional Context Reasoning} 
\label{sssec:stage02}
The emotional context reasoning stage aims to infer the emotional context of the dialogue by considering both the content of each utterance and the perceived emotions. 
We consider $N$ utterances in a conversation, denoted as $\{u_0, u_1, \dots, u_k, \dots u_N\}$, where $u_k$ represents the $k$-th utterance in the conversation history. 
Each utterance $u_k$ is associated with a corresponding ground-truth transcript $t_k$, speaker ID $s_k$, and textual emotion label $e_k$. 
We fine-tune the LLM with multiple PLoRA modules to predict the appropriate emotional state of the target utterance $u_c$ within the given dialogue history, where $c \in [2, N]$ represents the current turn. 
For this process, we generate the instruction embedding $I$ and the $k$-th utterance embedding $U_k$, corresponding to the $k$-th utterance $u_k$ in the dialogue history, where $k \in [0, c]$. 
These embeddings are generated from the EQ-former pre-trained in the first stage and the LLM's tokenizer for the LLM's input, as illustrated in Fig. \ref{fig:overview} (a) and (b). 

Each utterance embedding $U_k$ is formed by concatenating the prefix embedding $P_k$, emotion embedding $\hat{E}_k$, and transcript embedding $T_k$ along the temporal dimension. 
The prefix embedding $P_k$ and transcript embedding $T_k$ are text embeddings generated from the prefix text $p_k$ and the ground-truth transcript $t_k$ using the LLM's tokenizer. 
The prefix text $p_k$ indicates the order in which speaker $s_k$ is speaking. 
We utilize the Whisper decoder during inference to generate $T_k$ from speech. 
The emotion embedding $\hat{E}_k$, which represents the recognized emotional state of the $k$-th utterance, is derived from the EQ-former. 
For generating the current utterance embedding $U_c$, only $P_c$ and $T_c$ are used. 
The instruction embedding $I$, generated from the text instruction prompt by using the LLM's tokenizer, guides the model to predict the emotion and intensity for the current utterance, based on $U_{0:c}$ and $I$. 

To address the scarcity of emotional conversation datasets for emotional context reasoning, we adopt a pre-training strategy in the second stage. 
This strategy focuses on training the PLoRA-T module using only textual data, including emotion labels $e_k$, from DailyTalk \cite{10095751} and DailyDialog \cite{li-etal-2017-dailydialog}. 
In this pre-training, the emotion embedding $E_k$ is derived from the ground-truth emotion label $e_k$ in text form and converted into a text embedding using the LLM's tokenizer.  
By deriving emotion embeddings from text, the PLoRA-T module can be trained on large-scale textual datasets, allowing the LLM to better infer emotional context in conversations. 
During this pre-training process, we train only the PLoRA-T adapter. 
We then fine-tune the LLM on the DailyTalk dataset to predict the emotional state of target utterance, initializing it with the pre-trained PLoRA-T module from this pre-training stage, along with the PLoRA-E module and EQ-former trained in the first stage. 
This approach alleviates the limitation of insufficient emotional conversation datasets and enhances the model's ability for emotional context reasoning.
\subsubsection{Emotional Context-Aware Speech Synthesis} 
\label{sssec:stage03}
The emotional context-aware speech synthesis stage generates speech with appropriate emotional expressions based on the inferred emotional state. 
We use FastSpeech 2 \cite{ren2021fastspeech} as the backbone model, incorporating an emotion encoder and an intensity encoder to generate target speech that appropriately reflects the predicted emotion and intensity based on the emotional context, as depicted in the top-right panel of Fig. \ref{fig:overview}. The models for the emotional context reasoning stage and emotional context-aware speech synthesis are trained separately and concatenated during inference.
\section{Experiments}
\label{sec:experiment}
\subsection{Experimental Setup}
\label{sec:setup}
For the first stage, we use the DailyTalk \cite{10095751}, CREMA-D \cite{6849440}, EmoV-DB \cite{adigwe2018emotional}, IEMOCAP \cite{busso2008iemocap}, MEAD \cite{kaisiyuan2020mead}, and TESS \cite{dupuis2010toronto} datasets, with a total of 80.6 hours of speech. 
We selected data corresponding to seven emotion categories (happy, sad, surprise, angry, fear, disgust, and neutral) and three intensity levels (weak, medium, and strong) available in each dataset. 
For the second stage, we use the DailyDialog \cite{li-etal-2017-dailydialog} text dataset with 13,118 dialogues and the DailyTalk \cite{10095751} dataset, which contains 20 hours of speech across 2,541 dialogues derived from DailyDialog. 
We added three intensity levels from ECSS \cite{liu2024emotion} to the DailyTalk dataset, which originally did not include intensity labels. 
All audio samples are re-sampled at 16 kHz during both the text-emotion alignment and emotional context reasoning stages. 
To train the third stage, we use the DailyTalk \cite{10095751} dataset. 
The audio is converted to Mel-spectrogram using short-time-fourier transform with an FFT size of 1024, a hop size of 256, and a window size of 1024, applying an 80-bins mel filter. 
All audio samples in this stage are down-sampled at 22.05 kHz for speech synthesis, and the data was split into training, validation, and test sets with a ratio of 8:1:1. 
\subsection{Implementation Details}
\label{sec:detail}
We use the official pre-trained Whisper Large v3 \cite{pmlr-v202-radford23a} and Vicuna-7B LLM \cite{vicuna2023}, which is a LLaMA model \cite{touvron2023llama} fine-tuned to follow instructions using the Vicuna instruction set. 
For Q-former, we employ 25 trainable query tokens with a dimension of 768. 
Multiple PLoRA adapters are injected into the projection layers for all queries and values in every LLaMA self-attention layer, with a rank of 8 and a scaling factor of 4.0 for both modules. 
In the first and second stages, we apply the AdamW optimizer \cite{loshchilov2018decoupled} with $\beta_1=0.9$, $\beta_2=0.999$, weight decay of 0.05, and a cosine learning rate decay, peaking at $3\times10^{-5}$ with a linear warmup of 3k steps and a minimum learning rate of $1\times10^{-5}$. 
During training, only the parameters of the EQ-former and the PLoRA modules are updated, while the TLTR module is frozen in the second stage. 
In the third stage, the AdamW optimizer settings are $\beta_1=0.9$, $\beta_2=0.98$, and HiFi-GAN \cite{NEURIPS2020_c5d73680} is used as the vocoder. 
All models are trained for 180k steps in the first stage, 30k steps in the second stage, and 275k steps in the third stage, using 4 NVIDIA RTX A6000 GPUs. 
\subsection{Evaluation Metrics}
\label{sec:metric}
We conduct two mean opinion score (MOS) tests to subjectively evaluate the naturalness in dialogue context (N-DMOS) and the alignment of emotional expression with the dialogue's emotional context (E-DMOS). 
For MOS evaluation, we use Amazon MTurk to crowdsource 20 listeners, and the results are reported with a 95\% confidence interval. 
To evaluate whether synthesized speech contains appropriate emotional expressions, we use the emotion2vec plus large model \cite{ma-etal-2024-emotion2vec}, a speech emotion recognition foundation model fine-tuned on approximately 42,000 hours of emotional data, to measure emotional classification accuracy (ECA). 
To evaluate pronunciation accuracy, we calculate the word error rate (WER) and phoneme error rate (PER) using Whisper \cite{pmlr-v202-radford23a} and wav2vec 2.0 \cite{NEURIPS2020_92d1e1eb}, respectively. 
To evaluate prosody, we compute the pitch error (RMSE$_{f0}$), periodicity error (RMSE$_{p}$), F1 score of voiced/unvoiced classification (F1$_{v/uv}$), and the Mel-cepstral distortion (MCD). 
We conducted a duration prediction performance evaluation using the average absolute differences of the utterance duration (DDUR) \cite{8607053}. 
To evaluate the emotional context reasoning, we calculate the accuracy of emotion and intensity predictions using the weighted accuracy (WA), unweighted average accuracy (UA), and macro F1 score (F1). These values are reported as percentages.

\begin{table}[!t]
  \centering
  \caption{Evaluation and Ablation Study Results for Prediction of the Emotion and Intensity in Emotional Context Reasoning Stage.} \vspace{-0.15cm}
  \label{table_ablation}
   \resizebox{1.01\columnwidth}{!}{
      \begin{tabular}{l|cccc|ccc}
        \toprule
        \multicolumn{1}{c}{} & \multicolumn{4}{c}{\textbf{Emotion}} & \multicolumn{3}{c}{\textbf{Intensity}} \\
        \addlinespace[3pt]
        \textbf{Method}  & \textbf{ECA} & \textbf{WA} $(\uparrow)$ & \textbf{UA} $(\uparrow)$ & \textbf{F1} $(\uparrow)$ & \textbf{WA} $(\uparrow)$ & \textbf{UA} $(\uparrow)$ & \textbf{F1} $(\uparrow)$ \\
        \midrule
        ECSS \cite{liu2024emotion} & 55.72 & 43.51  & 15.06  & 13.66  & 60.38 & 33.33 & 25.10 \\
        JELLY & \textbf{58.60} & \textbf{78.54} & 59.09 & 60.17 & \textbf{77.21} & \textbf{52.78} & 51.61 \\
        \midrule
        w/o TLTR &  57.71 &  77.77 & \textbf{60.14} & 59.81 & 75.88 & 52.09 & 50.87 \\
        w/o Q-former & 55.38 & 47.01 & 14.81 & 10.43 & 59.62 & 36.10 & 32.46 \\
        w/o PLoRA modules & 57.05 & 77.77 & 59.24 & \textbf{62.52} & 76.33 & 51.71 & 50.83 \\
        w/o stage 1 & 57.38 & 76.99 & 57.60 & 58.12 & 74.00  & 52.03 & 49.80 \\
        w/o PT in stage 2 &  56.60 & 67.15 & 24.73 & 25.73 & 74.89 & 51.55 & \textbf{53.03} \\
        \bottomrule
      \end{tabular}
  }\vspace{-0.2cm}
\end{table}
\section{Results}
\label{sec:results}
We compared our JELLY framework with other CSS frameworks, including ECSS \cite{liu2024emotion} and a GRU-based approach \cite{9383460} as baselines. 
For ECSS, we used the official implementation for training. 
To verify that JELLY infers emotional context and generates contextually appropriate speech from speech alone, we also evaluated JELLY (speech-only), which performs inference using only speech through the EQ-former and Whisper. 
Additionally, we compared it with our TTS backbone model, FastSpeech 2 \cite{ren2021fastspeech}, without context modeling.
\subsection{Emotional Context-Aware Speech Synthesis}
\label{sec:result_1}
We conducted both subjective and objective evaluations to assess whether the generated speech appropriately reflects the emotional context. 
Table \ref{table_main} shows that JELLY outperforms the other baseline models across nearly all metrics, particularly in terms of N-DMOS, E-DMOS, ECA, MCD, and DDUR. 
The high E-DMOS and ECA scores validate that our framework generates natural speech with emotional expressions closely aligned with the emotional context. 
Specifically, the results show that JELLY, unlike other CSS baselines, effectively understands the emotional context and generates speech that is appropriate for the emotional context. 

The low WER and PER scores indicate that JELLY generates speech with reasonably accurate pronunciation, and the low DDUR score highlights JELLY's ability to model duration within the emotional context. 
Low error rates in terms of RMSE${_p}$ and MCD, along with the high scores in N-DMOS and F1$_{v/uv}$, demonstrate that our framework models prosody appropriate for conversational contexts. 

JELLY (speech-only) also demonstrates that our framework can generate emotionally context-aware speech without relying on transcripts or emotion labels during inference, as reflected in its high E-DMOS and ECA scores.
Notably, JELLY (speech-only) outperforms the comparison models in MCD, DDUR, and F1$_{v/uv}$, further showing that our framework, through its outstanding emotional context reasoning, generates speech with prosody well-suited to conversational contexts using only speech. 
\subsection{Emotional Context Reasoning}
\label{sec:result_2}
We conducted comparative experiments on WA, UA, and F1.
ECSS \cite{liu2024emotion} was used as the baseline, and its emotion and intensity predictors were employed for comparison with our framework. 
Table \ref{table_ablation} presents the results, demonstrating that JELLY outperformed the baseline across all metrics for both emotion and intensity predictions. 
Although ECSS exhibited inaccuracies in prediction, the prosody predictor in ECSS, which predicts prosody based on previous conversations, appears to compensate for these errors, as reflected in the high N-DMOS and low RMSE$_{f0}$ results shown in Table \ref{table_main}. 
JELLY infers emotional context from recognized emotions using the proposed EQ-former, even without ground truth emotion labels, leading to more accurate predictions of the target utterance's emotional state. 
\subsection{Ablation Study}
\label{sec:result_3}
As detailed in Table \ref{table_ablation}, we conducted ablation studies to verify the performance of each module in our proposed method.
We assessed the impact of TLTR by replacing it with the last hidden layer representations of Whisper encoder, which led to performance declines in ECA, WA, and F1 macro in the emotion category and across all intensity metrics. 
This emphasizes the role of TLTR in extracting emotion features, including detailed information such as emotion intensity, and improving emotional context reasoning. 

Next, removing Q-former and directly feeding emotion features from TLTR into the LLM through a projection layer led to a significant performance drop across all metrics, confirming that the Q-former plays a crucial role in bridging the modality gap between text and emotion, which is essential for the emotional context reasoning. 

When we replaced the multiple PLoRA modules with the standard LoRA adapter \cite{hu2022lora}, the decline in ECA and intensity metrics shows that the multiple PLoRA modules contribute to more effective modeling of intensity, ensuring alignment with the conversational context.

Additionally, we conducted an experiment where emotional inference was performed directly in the second stage, without the first stage. 
The decreases across all metrics indicate that EQ-former pre-trained in the first stage is effective for inferring emotional context and demonstrates the benefits of the value of using relatively available emotional speech data. 

Finally, skipping the pre-training of the PLoRA-T module in the second stage also led to lower performance across all metrics, highlighting the importance of pre-training with text data to address the scarcity of emotional conversational datasets and enhance the LLM's ability to reason the emotional context. 
\section{Conclusion}
\label{sec:conclusion}
In this paper, we proposed JELLY, a novel CSS framework that integrates joint emotion recognition and context reasoning with a large language model. 
We introduced the EQ-former as an emotion perception module for emotion-text alignment and designed a three-stage learning pipeline to mitigate the scarcity of emotional conversational datasets. Experimental results demonstrate that JELLY generates emotionally aligned and natural speech in conversational contexts, outperforming existing models—even without transcripts or emotion labels. 
By leveraging the reasoning capabilities of the LLM through multiple LoRA modules and a distinct pre-training strategy in the first and second stages, JELLY improves predictions of the target speaker’s emotional states and synthesizes speech that is more appropriate for the emotional context.
However, challenges remain in bridging the gap between CSS and real-world scenarios. 
In future work, we plan to address this by considering overlapped speech and larger speaker groups.
\vfill
\pagebreak

\balance
\bibliographystyle{IEEEbib}
\bibliography{refs}

\end{document}